\RequirePackage{fix-cm}
\documentclass[smallextended]{svjour3}       
\smartqed  
\usepackage{graphicx}
\usepackage{amsmath}
\usepackage{latexsym}
\usepackage{amsfonts}
\usepackage[normalem]{ulem}
\usepackage{soul}
\usepackage{array}
\usepackage{amssymb}
\usepackage{extarrows}
\usepackage{xcolor}
\usepackage{subfig}
\usepackage{wrapfig}
\usepackage{txfonts}
\usepackage{wasysym}
\usepackage{enumitem}
\usepackage{adjustbox}
\usepackage{ragged2e}
\usepackage{longtable}
\usepackage{changepage}
\usepackage{setspace}
\usepackage{hhline}
\usepackage{multicol}
\usepackage{tabto}
\usepackage{float}
\usepackage{multirow}
\usepackage{makecell}
\usepackage{mathptmx}      

\begin{document}

\title{Attention mechanisms and deep learning for machine vision: A survey of the state of the art}


\author{       Abdul Mueed Hafiz         	\and
		Shabir Ahmad Parah	\and		
        		Rouf Ul Alam Bhat 	
}


\institute{Abdul Mueed Hafiz \at
              Dept of Electronics \& Communication Engineering, Institute of Technology, University of 	  
	   Kashmir (Zakura Campus), Srinagar, J\&K, 190006 India\\
              Tel.: +91-7006474254\\
              \email{mueedhafiz@uok.edu.in}           
           \and
	 Shabir Ahmad Parah \at
            Department of Electronics and Instrumentation Technology, University of Kashmir (Main Campus), 
	 Srinagar, J\&K, 190006 India\\
	\and
	Rouf Ul Alam Bhat \at
              Dept of Electronics \& Communication Engineering, Institute of Technology, University of 	  
	   Kashmir (Zakura Campus), Srinagar, J\&K, 190006 India\\
}

\date{Received: date / Accepted: date}

\maketitle

\begin{abstract}
With the advent of state of the art nature-inspired pure attention based models i.e. transformers, and their success in natural language processing (NLP), their extension to machine vision (MV) tasks was inevitable and much felt. Subsequently, vision transformers (ViTs) were introduced which are giving quite a challenge to the established deep learning based machine vision techniques. However, pure attention based models/architectures like transformers require huge data, large training times and large computational resources. Some recent works suggest that combinations of these two varied fields can prove to build systems which have the advantages of both these fields. Accordingly, this state of the art survey paper is introduced which hopefully will help readers get useful information about this interesting and potential research area. A gentle introduction to attention mechanisms is given, followed by a discussion of the popular attention based deep architectures. Subsequently, the major categories of the intersection of attention mechanisms and deep learning for machine vision (MV) based are discussed. Afterwards, the major algorithms, issues and trends within the scope of the paper are discussed.

\end{abstract}

\section{Introduction}

Recently attention-based mechanisms like transformers 
\cite{tv_1} have been successfully applied to various machine 
vision tasks by using them as vision transformers (ViTs) \cite{tv_11} in image 
recognition \cite{tv_12}, object detection 
\cite{tv_13,tv_14}, segmentation\cite{tv_15}, image super-resolution 
\cite{tv_16}, video understanding 
\cite{tv_17,tv_18}, image generation 
\cite{tv_19}, text-image synthesis 
\cite{tv_20} and visual question answering 
\cite{tv_21,tv_22}, among others 
\cite{tv_23,tv_24,tv_25,tv_26} achieving at par as well 
as even better results as compared to the established CNN models 
\cite{tv}. However, transformers have various issues like being 'data-hungry' and 
requiring large training times. Deep learning \cite{dl_main,dl_book,dl_review} based convolutional neural
networks (CNNs) \cite{cnn1,cnn2} on the other hand do not have such problems significantly. Accordingly, techniques have emerged which are at the 
intersection of pure attention based models and the established pure 
CNNs which have best of the both features. Machine vision (MV) has also 
benefitted from this merger of the two important vision models viz. ViTs 
and CNNs. In the this section we will discuss the source of power 
of ViTs and transformers in general i.e. attention and its types \cite{tv} briefly for the 
readers to have an idea of the new type of machine vision (MV) models 
i.e. ViTs. 

\subsection{Self-attention}

For a given a sequence of elements, the self-attention process gives a 
measurable estimate of the relevance of one element others. For example, 
which elements like words can come together in a sequence like a 
sentence. The self-attention process is an important unit of 
attention-based models like transformers, that models the dependencies 
among all elements of the sequence for formal/structured prediction 
applications. Plainly stated, a self-attention model layer assigns a 
value to every element in a structure/sequence by combining information 
globally from the input vector/sequence. 

Denoting a sequence of \textit{n} entities (\textbf{x}\textsubscript{1}, \textbf{x}\textsubscript{2}, $ \ldots $  \textbf{x}\textsubscript{n}) by  \( \textbf{X} \in \mathbb{R}^{n \times d} \) , \textit{d} being the dimension which embeds dependency of every element. The purpose of self-attention is capturing the dependency between all \textit{n} elements after encoding every element inside the overall contextual knowledge. This process is achieved\  by the definition of 3 weight matrices which have to be learnt for transforming: Queries  \(  \left( \textbf{W}^{Q} \in \mathbb{R}^{n \times d_{q}} \right)  \), Keys  \(  \left( \textbf{W}^{K} \in \mathbb{R}^{n \times d_{k}} \right)  \)  and Values  \(  \left( \textbf{W}^{V} \in \mathbb{R}^{n \times d_{v}} \right)  \). First the input vector \textbf{X} is projected to the 3 weight matrices for obtaining  \( \textbf{Q}=\textbf{XW}^{Q} \),  \( \textbf{K}=\textbf{XW}^{K} \)  and  \( \textbf{V}=\textbf{XW}^{V} \). The output  \( \textbf{Z} \in \mathbb{R}^{n \times d_{v}} \)  in the self-attention layer is next expressed as,
\begin{equation}
 Z=\textbf{softmax} \left( \frac{\textbf{QK}^{T}}{\sqrt[]{d_{q}}} \right) \textbf{V}. 
\end{equation}

For a certain element in the vector/sequence, the self-attention mechanism fundamentally finds the dot product of query with all the keys, this product being subsequently normalized by the softmax function for obtaining the attention-map scores. Every element now assumes the value of the weighted summation for all elements inside the vector/sequence, wherein all weights are equal to the attention map scores.

\subsection{Masked self-attention}

\begin{justify}
The self-attention layer applies to every element/entity. For the transformer \cite{tv_1}  having been trained for prediction of the next entity in the vector/sequence, the self-attention units inside the decoder are then masked for prevention of their application to the entities coming in future. This technique is achieved by calculating the element-wise product with a mask  \( \textbf{M} \in \mathbb{R}^{n \times n} \) , where \textbf{M} is the upper triangular matrix. Thus masked self-attention is calculated as, 
\end{justify}

\begin{equation}
\textbf{softmax} \left( \frac{\textbf{QK}^{T}}{\sqrt[]{d_{q}}} {\circ} ~\textbf{M} \right),
\end{equation}

where  \(  {\circ}  \)  is the Hadamard product. During prediction of an element in the vector/sequence, the attention map scores of the future elements are set to 0 in the masked self-attention. 

\subsection{Multi-head attention}

\begin{justify}
For encapsulation of various complicated dependencies between various elements / entities in the vector/sequence, the multi-head attention process consists of multiple self-attention units with \textit{h =} 8 inside the original transformer architecture \cite{tv_1}. Every unit contains its own learnable weight-matrices  \(  \left\{ \textbf{W}^{Q_{i}},\textbf{W}^{K_{i}},\textbf{W}^{V_{i}} \right\} \), where \textit{i = 0, 1, 2, ... (h - 1)}. For a particular input \textbf{X}, outputs of \textit{h} self-attention units in the multi-head attention process are combined into one matrix \( [\textbf{Z}\textsubscript{0}, \textbf{Z}\textsubscript{1}, $ \ldots $  \textbf{Z}\textsubscript{h$-$ 1}]  \in \mathbb{R}^{n \times h \times d_{v}} \) and are subsequently projected to another weight matrix \textbf{W} $ \in $   \( \mathbb{R}^{h \cdot d_{v} \times d}. \)  
\end{justify}

\begin{figure}
\includegraphics[width=12 cm]{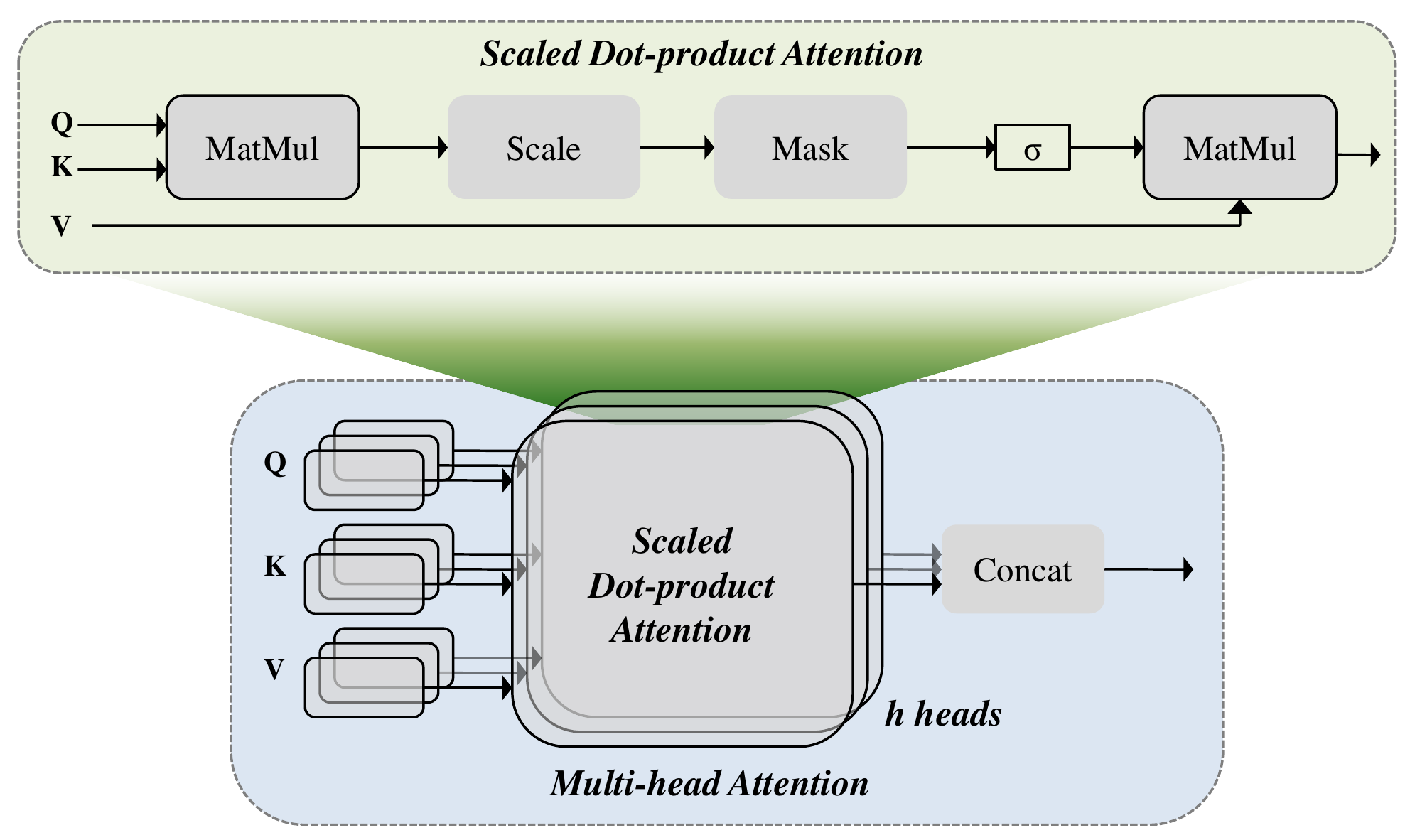}
\caption{Illustration of various attention mechanisms \cite{tv}}
\label{fig1}
\end{figure}

The notable difference of the self-attention process with the 
convolutional operation is that every weight is dynamically computed as 
against static weights which remain fixed for various inputs as for 
convolution. Also that the self-attention process is invariable to 
permutation and change for different number of inputs with the result 
that it has a convenient operation over irregularity as against the 
convolutional operator which needs a grid array. See \ref{fig1} for 
illustration of these concepts. 

\section{Attention based deep learning architectures}

\begin{figure}
\includegraphics[width=12 cm]{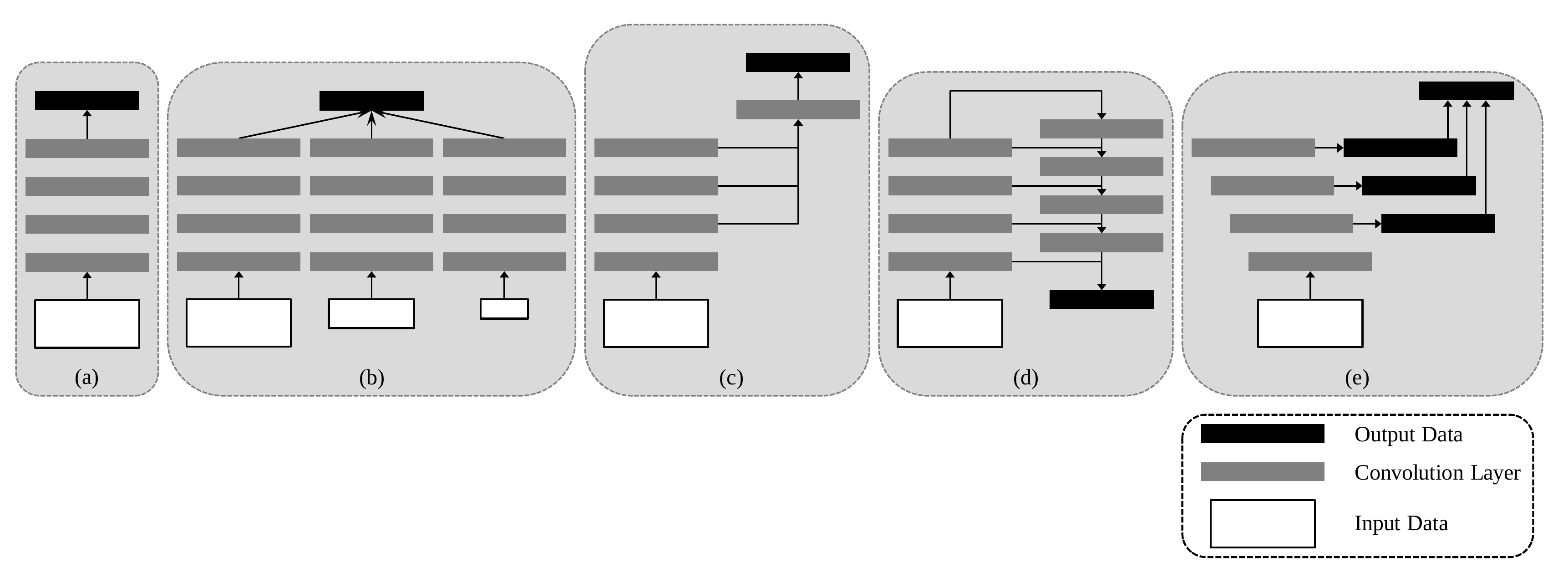}
\caption{(a)-(c) Depiction of 3 common classes of deep learning configurations used in attention prediction: (a) single-channel model configuration, (b) multi-channel model configuration with multi-scale input data, and (c) skip-layer model configuration. (d) bottom-up/top-down model configuration used in attention-based object segmentation and instance segmentation. (e) modified skip-layer model using multi-scale attention information in a single network. \cite{DeepVisualAttentionPred}}
\label{fig2}
\end{figure}

In this section, some common deep learning architectures of deep 
attention models are discussed \cite{DeepVisualAttentionPred} and a graphical illustration 
is presented in \ref{fig2}. The architectures of the prevalent deep 
attention based models are categorized into the following important 
classes as given below: 

\newcounter{numberedCntE}
\begin{enumerate}
\item Single channel model
\item Multi-channel model feeding on multi-scale data 
\item Skip-layer model
\item Bottom-up/ top-down model
\item Skip-layer model with multi-scale saliency single network
\setcounter{numberedCntE}{\theenumi}
\end{enumerate}

\subsection{Single-channel model}

As demonstrated in Figure 2(a), the single channel model is the 
predominant configuration of various CNN-based attention models also 
being used by many attention-based works 
\cite{a51,a14,a16,a15}. Almost all the other types of CNN 
configurations can be considered as variants of the single channel 
model. It has been demonstrated that attention cues on various levels 
and scales are vital for attention \cite{a28}. Using 
multi-scale features of CNNs into attention-based models is an obvious 
choice. In the next type of single channel model, namely multi-channel 
model, the changes are done along this line. 

\subsection{Multi-channel model}

Some implementations of this model include 
\cite{a52,a57,a32,a4}. The basic concept in the 
multi-channel model is shown in Figure 2(b). This type of model learns 
multi-scale attention information by training multiple models with 
multi-scale data inputs. The multiple model channels are in parallel and 
can have varying configurations with different scales. As shown in 
\cite{a58}, input data is fed via multiple channels 
simultaneously, and then the features from different channels are fused 
and fed into a unified output layer for producing the final attention 
map. We observe that in the multi-channel model, multi-scale learning 
takes place outside the individual models. In the next configuration 
discussed, the multi-scale learning is inside the model, and this is 
achieved by combining feature maps from various convolutional layer 
hierarchies.

\subsection{Skip-layer model}

A common skip-layer model is shown in Figure 2(c) being used in 
\cite{a48,a50,a59}. Instead of learning from many parallel channels on multiple-scale images, the skip-layer model learns 
multi-scale feature maps inside a primary channel. Multi-scale outputs 
are learned from various layers with increasingly larger reception 
fields and down-sampling ratios. Next, these outputs are fused for 
outputting final attention map. 

\subsection{Bottom-up/top-down model}

This relatively newer model configuration called top-down/bottom-up 
model has been used in attention-based object segmentation 
\cite{a60} and also in instance segmentation 
\cite{a61,hafizmmir,mmir_review}. The architecture of the model is shown in 
Figure 2(d), wherein segmentation feature maps are first obtained by 
common bottom-up convolution techniques, and next a top-down refinement 
is done for fusing the data from deep to shallow layers into the mask. 
The main motivation behind this configuration is to produce 
high-fidelity segmentation masks because deep CNN layers lose fine image 
detail. The bottom-up/top-down model is like a type of skip-layer model 
since different layers are connected to each other. 

\subsection{Skip-layer Model with Multi-scale Saliency Single Network}

This model \cite{DeepVisualAttentionPred} shown in Fig. 
2(e), is inspired by the model in \cite{a58} and the deeply-supervised 
model in \cite{a22}. The model uses multi-scale and 
multi-level attention-based information from various layers, and learns 
via the deeply supervised technique. An important difference between 
this model and the previous models is that the former provides combined 
straightforward supervision of the hidden layers instead of the common 
approach of supervising only the last output layer and then propagating 
the supervised output back to the previous layers. It uses the merit of 
the skip-layer model (Figure 2(c)) which does not learn from multiple 
model channels with multi-scale input data. Also, it is lighter than the 
multi-channel model (Figure 2(b)) and bottom-up/top-down model (Figure 
2(d)). It has been found that the bottom-up/top-down model faces 
training difficulties while as the deeply supervised model shows high 
training efficiency.

In the next section we turn to the categorization of various techniques 
of attention mechanisms and deep learning in machine vision, and discuss 
each category in detail.

\section{Attention and deep learning in machine vision: Broad categories}

In this section, we discuss category-wise the various techniques of 
attention mechanisms and deep learning applied to machine vision. Three 
broad categories are:

\newcounter{numberedCntF}
\begin{enumerate}
\item Attention-based CNNs
\item CNN transformer pipelines
\item Hybrid transformers 
\setcounter{numberedCntF}{\theenumi}
\end{enumerate}
These categories are discussed in the following sub-sections one by one. 
First we discuss attention-based CNNs in the following subsection.

\subsection{Attention-based CNNs}

Recently attention mechanisms have been applied in deep learning for 
machine vision applications, e.g. object detection 
\cite{ac1_3,ac1_31,ac1_28}, image captioning 
\cite{ac1_35,ac1_39,ac1_2} and action recognition 
\cite{ac1_30}. The central idea of the attention 
mechanisms is locating the most salient components of the feature 
maps in convolutional neural networks (CNNs) in a manner that the 
redundancy is removed for machine vision applications. Generally, 
attention is embedded in the CNN by using attention maps. Particularly 
the attention-based maps in 
\cite{ac1_31,ac1_28,ac1_35,ac1_30} yield in a self 
learned manner having other information with weak supervision of the 
attention maps. Other techniques cited in literature 
\cite{ac1_39,ac1_37} proceed by utilization of human 
attention data or guidance of the CNNs by focusing on the regions of 
interest (ROIs). In the following subsections, we proceed with 
discussing some noteworthy techniques in the general area of machine 
vision which use attention-based CNNs e.g. those used in image 
classification/retrieval, object detection, sign language recognition, 
denoising and facial expression recognition.

\subsubsection{Image classification/retrieval and object detection}

\begin{figure}
\includegraphics[width=12 cm]{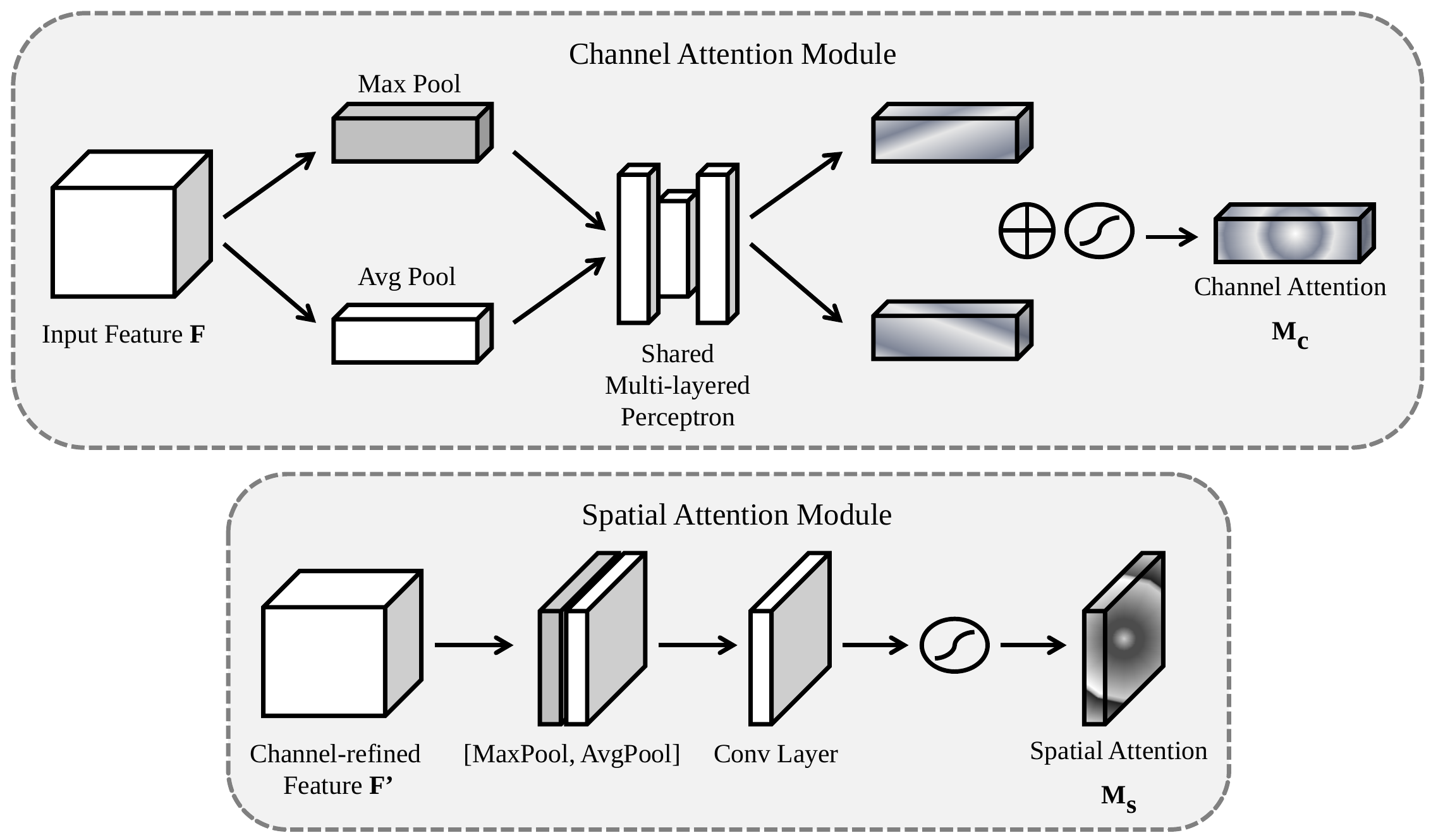}
\caption{Illustration of both attention sub-modules in CBAM \cite{ac2}. As shown, the channel-wise sub-module utilizes the max-pooling of the feature output as well as the average-pooling of the feature output with the help of a shared network. On the other hand, the spatial-wise sub-module uses two identical feature outputs by pooling them along their channel axes and then forwarding them to the convolutional layer. \cite{ac2}}
\label{fig3}
\end{figure}

It is well established that attention contributes to human perception in 
an important manner \cite{ac2_2,ac2_24,ac2_25}. One 
important characteristic of a human vision system is that it does not 
attempt to address the whole visual scene at one go. Instead, in the 
same, a sequence of partial glimpses is exploited and focusing is done 
selectively on various parts for capturing the visual structure in a 
better manner \cite{ac2_26}. Recently, several attempts 
have been made \cite{ac2_27,ac2_28} for incorporation of 
attention processing mechanisms in order to improve the classification 
accuracy of CNNs on large scale classification tasks. Wang et al. 
\cite{ac2_27} have proposed a residual attention network 
having an encoder-decoder style attention mechanism unit. By refinement 
of the features, the network gives good accuracy as well as shows 
robustness to noise. Without directly computing the three dimensional 
attention map, the process is decomposed such that it learns 
channel-attention and spatial-attention exclusively. The exclusive 
attention map generation technique for 3D features is computationally 
inexpensive and parameter restricted, and hence can be used as a plug 
and play unit for existing CNN networks. In their work 
\cite{ac2_28}, the authors have introduced a compact unit 
for exploitation of the relationship between various channels. In this 
'Squeeze and Excitation' unit, the authors have used global 
average-pooling of feature maps for computation of each channel's 
attention. However, the authors of \cite{ac2} show that the 
features used in \cite{ac2_28} are suboptimal for 
inferring fine-channel attention. Accordingly the authors of 
\cite{ac2} use max-pooled feature maps also. According to 
\cite{ac2} in \cite{ac2_28} spatial attention 
is missed which contributes in an important manner to deciding the 
focusing region as brought out in \cite{ac2_29}. The authors of
\cite{ac2} thus proposed the convolutional block attention 
module (CBAM) for exploitation of both the spatial as well as 
channel-wise attention with the help of a robust network and proceed to 
verify that exploitation of both these mechanisms is better than use of 
only the channel-wise attention mechanism \cite{ac2_28} by 
using it for image classification in ImageNet-1K dataset 
\cite{ac2_imagenet}. The authors of \cite{ac2} experimentally 
demonstrate that their module is effective also in object 
detection tasks using two popular datasets viz. MS-COCO 
\cite{ac2_coco} and VOC \cite{ac2_voc}. They 
achieve impressive results by inserting their module in the pre-existing 
one-shot object detector \cite{ac2_30} in the VOC-2007 
testing set. \ref{fig3} shows the CBAM for both channel and 
spatial-attention processes. Here we attempt to briefly explain
the attention mechanism in CBAM.

\begin{justify}
For a given input feature map  \( \textbf{F} \in \mathbb{R}^{C \times H \times W} \) \textbf{, }CBAM \cite{ac2}  produces a one-dimensional attention map  \( \textbf{M}_{c} \in \mathbb{R}^{C \times 1 \times 1} \) \textbf{ }and a two-dimensional spatial attention map  \( \textbf{M}_{s} \in \mathbb{R}^{1 \times H \times W} \) \textbf{ }as shown in Figure 3. This attention mechanism operation can be put as:
\end{justify}

\begin{equation}
\textbf{F}^{'}=\textbf{M}_{c} \left( \textbf{F} \right)  \bigotimes \textbf{F},
\end{equation}

\begin{equation}
\textbf{F}^{''}=\textbf{M}_{s} \left( \textbf{F}' \right)  \bigotimes \textbf{F}'.
\end{equation}

\begin{justify}
where  \( \bigotimes \) \textbf{ }is the multiplication operator for elements.
\end{justify}
\begin{justify}
Channel attention is mathematically computed as follows:
\end{justify}

\begin{multline}
\textbf{M}_{c} \left( \textbf{F} \right) = \sigma  \left( MLP \left( AvgPool \left(\textbf{F} \right)  \right) +MLP \left( MaxPool \left( \textbf{F} \right)  \right)  \right) \\
= \sigma  \left( \textbf{W}_{1} \left( \textbf{W}_{0} \left( \textbf{F}_{\mathrm{avg}}^{c} \right)  \right) +\textbf{W}_{1} \left( \textbf{W}_{0} \left( \textbf{F}_{\mathrm{\max }}^{c} \right)  \right)  \right)
\end{multline}

\begin{justify}
where  \(  \sigma  \)  is the sigmoid function,  \( \textbf{W}_{0}  \in \mathbb{R}^{C/r \times C} \)  and  \( \textbf{W}_{1}  \in \mathbb{R}^{C \times C/r} \) . The Multi-layer Perceptron (MLP) weights,  \( \textbf{W}_{0} \)  and  \( \textbf{W}_{1} \) , are shared for both the inputs.  \( \textbf{W}_{0}~ \) comes after the ReLU activation function.
\end{justify}

\begin{equation}
\textbf{M}_{s} \left( \textbf{F} \right) = \sigma  \left( f^{7 \times 7} \left(  \left[ AvgPool \left( \textbf{F} \right) ;MaxPool \left( \textbf{F} \right)  \right]  \right)  \right) = \sigma  \left( f^{7 \times 7} \left(  \left[ \textbf{F}_{\mathrm{avg}}^{s};\textbf{F}_{\mathrm{\max }}^{s} \right]  \right)  \right)
\end{equation}

\begin{figure}
\includegraphics[width=12 cm]{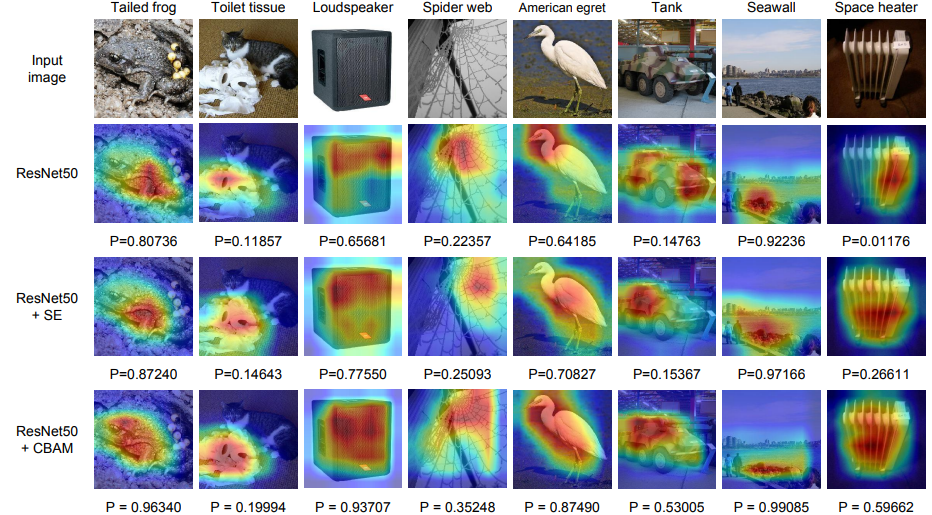}
\caption{Heat map visualizations using Grad-CAM \cite{ac2_18}. The visualizations are shown for those of the CBAM-fitted CNNs, viz. \{\textbf{ResNet50 + CBAM}\}, baseline \{\textbf{ResNet50}\} \cite{ac2_5}, and Squeeze and Excitation method \cite{ac2_28} (SE)-integrated architecture \{\textbf{ResNet50 + SE}\}. Grad-CAM visualization has been obtained with feature maps of last conv layer outputs. The GT label has been shown on top of every image, where P is the softmax score of every network for the GT category. (Reproduced by permission from publisher of \cite{ac2})}
\label{fig4}
\end{figure}

\begin{justify}
where  \(  \sigma  \)  is the sigmoid function and  \( f^{7 \times 7} \)  is the convolutional operator with a  \( 7 \times 7 \)  filter.
\end{justify}
The authors of \cite{ac2} have used their technique for 
both image classification/retrieval on the ImageNet-1K dataset and 
object detection on both MS-COCO and VOC2007 datasets. The 
results obtained using their CBAM integrated networks outperform other 
contemporary networks. They also have demonstrated the superiority of 
their technique as compared to others also via grad-CAM 
\cite{ac2_18} visualizations obtained on images from 
ImageNet validation set. \ref{fig4} shows the same.

Another novel and related work in the area of image 
classification/retrieval by using attention-based CNNs is given by the 
authors of \cite{ac1} for glaucoma detection from the area 
of medical image analysis \cite{hafizmedical}. They call their network attention-based CNN 
for glaucoma detection AG-CNN. It includes a novel 
attention-prediction subnet along with other subnets. They achieve 'end 
to end' training on an attention-based CNN architecture by supervision 
of the training through 3 separate loss functions based on: i) 
attention-prediction, ii) feature-visualization and iii) 
glaucoma-classification. Based on the work of authors in 
\cite{ac1_16}, the authors use the Kullback Leibler (KL) 
divergence function as an equivalent of the nature-inspired 
attention-loss $Loss_{a}$ given by:

\begin{equation}
\mathrm{Loss}_{a}=\frac{1}{I \cdot J} \sum _{i=1}^{I} \sum _{j=1}^{J}A_{ij}\log  \left( \frac{A_{ij}}{\hat{A}_{ij}} \right)
\end{equation}

\begin{justify}
where  \( \hat{A} \)  (with its elements  \( \hat{A}_{ij} \in  \left[ 0,1 \right]  \) ) is the attention map, and, \textit{I} and \textit{J} are the attention-map length and width respectively. By incorporating these novel features, the authors of \cite{ac1}  demonstrate that their proposed AG-CNN technique significantly improves the state of the art in glaucoma detection.
\end{justify}

For more interesting techniques on image classification/retrieval using 
attention-based CNNs the readers may refer to some of the recent 
outstanding works in this area as given in 
\cite{pb,hic}, etc.

\subsubsection{Sign Language Recognition}

Sign language recognition (SLR) is a valuable and challenging research 
area in machine vision related multimedia field. Conventionally, SLR 
relies on hand-crafted features with low performance. In their novel 
work \cite{sl}, the authors propose to use attention based 
3D CNNs for SLR. Their model has 2 advantages. First, it learns spatial 
and temporal features from video frames without any pre-processing or 
prior knowledge. Attention mechanisms help the model to select the 
clues. During training for capturing the features, spatial attention is 
used in the model for focusing on the ROIs. After this, temporal 
attention is used for selection of the important motions for determining 
the action-class. Their method has been benchmarked on a self-made large 
Chinese SL dataset having 500 classes, and also on the ChaLearn14 
benchmark \cite{sl_6}. The authors demonstrate that their 
technique outperforms other state of the art techniques on the datasets 
used. We discuss this interesting technique in more detail below.
\begin{justify}
The spatial attention map is calculated as follows. They use an attention-based mask for denotation of the value of each image pixel. Let  \( x_{i,k} \in \mathbb{R}^{2} \)  denote the position of a viewpoint \textit{k} in an image \textit{i}, the value of the location  \( p \in \mathbb{R}^{2} \)  inside the attention map  \( M_{i,k} \in \mathbb{R}^{w \times h} \)  for \textit{k} is given by:
\end{justify}

\begin{equation}
M_{i,k} \left( p \right) =\exp  \left( -\frac{ \Vert p-x_{i,k} \Vert _{2}^{2}}{ \sigma } \right) ,
\end{equation}

\begin{justify}
where  \(  \sigma  \)  is experimentally chosen, and \textit{w} and \textit{h} are image dimensions. The attention mask is formed by aggregating the peaks of various viewpoints obtained previously with the help of a max operator,
\end{justify}

\begin{equation}
M_{i} \left( p \right) =\mathop{\max }_{k}M_{i,k} \left( p \right).
\end{equation}

\begin{justify}
Consequently the \textit{i\textsuperscript{th}} attention weighed image \textit{I\textsubscript{i}} is the element-wise product given by,
\end{justify}

\begin{equation}
I_{i} \left( p \right) = I_{i} \left( p \right)  \times M_{i} \left( p \right).
\end{equation}

\begin{justify}
Based on the video feature obtained above, the use a Support Vector Machine (SVM) based classifier \cite{sl_18} for classification by clubbing it to another temporal attention-based pipeline. As done earlier in \cite{sl_43}, the features are fed to a bi-directional LSTM for generation of an attention vector  \( s \in \mathbb{R}^{8192} \) . The features are also fed to a one-layer MLP which gives the hidden vector  \( H= \{ h_{1},h_{2}, \ldots ,h_{n} \}  \) ,  \( h_{i} \in \mathbb{R}^{8192} \) . This vector is an integration of the sequence of clip features by attention pooling. This technique measures the value of each clip feature by determining its relation with the attention vector s. Finally, they combine the video and trajectory features and use softmax based classification. Although an effective technique, the authors still admit that the work focuses on isolated SLR. For dealing with continuous SLR, which translates a clip into a sentence, RNN based methods are going to give results as admitted by the authors of the above work, and they want to work in that direction.
\end{justify}

\subsubsection{Image denoising}

Image denoising is a low-level machine vision (MV) task. Deep CNNs are 
quite popular in low-level MV. Research has been done to improve the 
performance in the area by using very deep networks. However, as the 
network depth increases, the effects of the shallow layers on deep 
layers decrease. Accordingly the authors of \cite{ac4} have 
proposed an attention-based denoising CNN named ADNet\textit{ }
featuring an attention block (AB). The AB has been used for fine 
extraction of the noise data hidden in complex backgrounds. This 
technique has been proved by the authors of \cite{ac4} to 
be very effective for denoising images with complex noise e.g. real 
noise-induced images. Various experiments demonstrate that ADNet 
delivers very good performance for 3 tasks viz. denoising of synthetic 
images, denoising of real noisy images, and also blind denoising. Here, 
the AB guides the previous network section by using the current network 
section in order to learn the noise nature. This is particularly useful 
for unknown images having noise, i.e. real noisy images and blind 
denoising. The AB uses 2 successive steps for implementation of its 
attention mechanism. First a 1x1 convolution is done on the output from 
the 17$^{th}$ CNN layer output in order to compress the feature map into 
a weight vector for adjustment of the previous section. Next, the 
weights thus obtained are used to multiply the feature map output of the 
16$^{th}$ CNN layer for extraction of more refined noise feature 
maps. It should be noted that inspired by this novel effort more complex 
attention mechanisms can be used along with more dedicated 'denoising' 
deep CNNs. The code of ADNet is available at: \textcolor{red} {https://github.com/hellloxiaotian/ADNet}. 

\subsubsection{Facial expression recognition}

One hot topic in Machine Vision (MV) is facial expression recognition 
(FER) which can be used in various MV fields like human computer 
interaction (HCI), affective computing, etc. In their work 
\cite{ac31}, the authors have proposed an end to end CNN 
network featuring an attention mechanism for auto FER. It has 4 main 
parts viz. feature extraction unit, attention unit, reconstruction unit 
and classification unit. The attention mechanism incorporated guides the 
CNN for paying more attention to important features extracted from 
earlier unit. The authors have combined their LBP features and their 
attention mechanism for enhancing the attention mechanism for obtaining 
better performance. They have applied their technique to their own 
dataset and 4 others, i.e., JAFFE \cite{ac31_13}, CK+ 
\cite{ac31_11}, FER2013 \cite{ac31_39} and 
Oulu-CASIA \cite{ac31_25}, and have experimentally 
demonstrated that their technique performs better than other 
contemporary techniques. The attention mechanism used in the work has 
been proved to be valuable in pixelwise MV tasks. Their attention unit 
consists of two branches. The first is used to obtain feature map 
\textit{F$_{p}$}, and the second combines the LBP feature maps for 
obtaining the attention maps \textit{F$_{m}$}. In the next step, 
the element wise multiplication is done for the attention maps 
\textit{F$_{m}$} and the feature maps \textit{F$_{p}$} to 
obtain the final feature maps \textit{F$_{m}$} as: 

\begin{equation}
F_{final}=F_{p}F_{m}
\end{equation}

\begin{justify}
Supposing that input of previous layer in the second branch is \textit{f\textsubscript{m}}, then the attention maps \textit{F\textsubscript{m}} are given by: 
\end{justify}

\begin{equation}
 F_{m}=sigmoid \left( Wf_{m}+b \right) 
\end{equation}

where w and b are denotations for weights and bias of conv layer, 
respectively. The technique is suitable for 2D images and its 
architecture needs to be modified to extend its application to video, 3D 
facial data, depth-image data. The authors also state that they are 
considering using more robust and efficient machine learning (ML) 
techniques for enhancement of the architecture.

In another valuable work in the area of FER given in 
\cite{ac32},the authors state that in spite of the fact 
that conventional FER systems are almost perfect for analyzing 
constrained poses however they cannot perform well for partially 
occluded poses which are common in the real world. Accordingly, they 
have proposed an attention-based CNN (ACNN) for perception of facial 
occlusion part which focuses on the highly discriminative unoccluded 
parts. Learning in their model is end to end. For various Regions of 
Interest (ROIs), they have introduced two types of ACNN viz. patch based 
type and global-local based type. The first type uses attention only for 
local patches in face regions. The second type combines local features 
at the patch level with global features at the imagelevel. Evaluation is 
done on their own face expression dataset having in-the-wild occlusions, 
2 of the largest in-the-wild face expression datasets i.e. RAF-DB 
\cite{ac32_4} and AffectNet \cite{ac32_5} 
and many other datasets. They show experimentally that using ACNNs 
improves the FER performance wherein the ACNNs shift attention from 
occluded facial regions to others which are not. They also show that 
their ACNN outperforms other state of the art techniques on several 
important FER datasets. However, the technique relies on landmarks. The 
authors intend to address this issue, as according them, ACNNs rely on 
face landmark localization units. Hence ACNNs have to be made more 
robust for generation of attention maps without landmarks, and this is 
an open area for research.

In the next sub-section, we turn to another important category of 
techniques of attention mechanisms and deep learning in machine vision, 
namely CNN transformer pipelines.

\subsection{CNN transformer Pipelines}

In this sub-section, we discuss another important category of techniques 
of attention and deep learning in machine vision, namely the CNN 
transformer pipeline. Here a CNN is used to feed feature maps to a 
transformer, and acts like a teacher to the transformer, as will be 
discussed. The notable works falling under this category have been 
discussed below for each area of machine vision (MV).

\subsubsection{Image recognition}

Transformers are 'data-hungry' in nature. For example a large-scale 
dataset like ImageNet \cite{ac2_imagenet} is not 
sufficient to train a vision transformer from scratch. To address this 
issue, the work in \cite{tv_12} proposes to distill 
information from a teacher CNN to a student transformer, in turn 
allowing training of the transformer only on ImageNet sans additional 
data. The data-efficient image transformer (DeiT) 
\cite{tv_12} is a first in large scale image 
classification/retrieval without using a large-scale dataset like JFT 
\cite{tv_39}. DeiT shows that transformers (requiring very 
large amounts of training data) can also be trained successfully on 
medium-sized datasets (e.g., 1.2M images as against 100M+ images used in 
ViT \cite{tv_11}) with shorter training time. An important 
contribution of DeiT is its novel native distillation technique 
\cite{tv_78} which uses a teacher CNN (RegNetY-16GF 
\cite{tv_79}) whose outputs are fed to the transformer for 
training. The feature map outputs from the teacher CNN help the 
transformer (DeiT) in effectively finding important representations in 
input data images. The representations learned by DeiT are as good 
as top-performing CNNs like EfficientNet \cite{tv_80} and 
also are efficiently applicable to various downstream image recognition 
tasks.

\subsubsection{Object detection}

\begin{figure}
\includegraphics[width=12 cm]{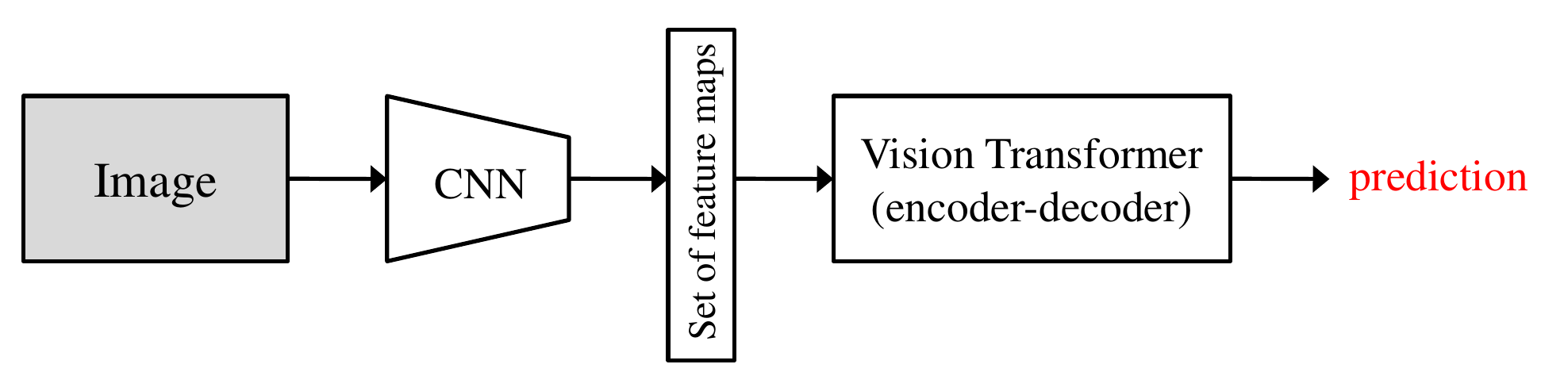}
\caption{Overview of the DETR pipeline. \cite{tv_13}}
\label{fig5}
\end{figure}

Like image classification/retrieval, transformers can be applied to 
image feature-map sets obtained from CNNs for precise object detection 
which involves prediction of object bounding boxes (BBoxes) and their 
corresponding category labels. In DETR \cite{tv_13}, given 
spatial features obtained from a CNN backbone, the transformer encoder 
flattens the spatial axes along a single axis as shown in \ref{fig5} which 
is feature map flattening from 3D to 1D. A sequence of features $
(d\times n)$ is obtained with $d=$ feature dimension, and $n=h\times 
w$ ($[h,w]$ being the size of the feature map). Next, the 1D 
flattened features are encoded and decoded by the multi-head 
self-attention units as given in the work of \cite{tv_1}. 

\subsubsection{Multi-modal machine vision tasks}

The machine vision (MV) tasks in this category include vision-language 
tasks like visual question-answering (VQA) \cite{tv_135}, 
visual commonsense-reasoning (VSR) \cite{tv_136}, 
crossmodal retrieval \cite{tv_137} and image-captioning 
\cite{tv_138}. There is a body of work for these areas 
within the scope of this paper, and the notable works have been 
mentioned here. In their work \cite{tv_22}, the authors 
propose VL-BERT \cite{tv_22}, one such technique for learning features which can be 
generalized to multi-modal MV downstream tasks like VSR and VQA. This 
technique involves aligning both visual as well as linguistic cues in 
order for learning compositely and effectively. For this, 
\cite{tv_22} uses the BERT (Bidirectional encoder 
representations from transformers) \cite{tv_3} 
architecture, and feeds it the features obtained from both visual and language 
domains. The language-features are the tokens in the input text 
sequences and the visual-features are the ROIs obtained from the input 
image by using a standard faster R-CNN model \cite{tv_83}. 
Their performance on various multi-modal MV tasks shows the advantage of the proposed technique 
over conventional 'language only' pre-training as done in the BERT 
\cite{tv_3}.

\subsubsection{Video understanding}

Videos which are audiovisual data are abundantly found. In spite of 
this, the contemporary techniques tend to learn from short videos (up to 
few seconds) allowing them to interpret usually short-range 
relationships \cite{tv_1,tv_29}. Long-range relationship 
learning is needed in different uni-modal and multi-modal MV tasks like 
activity recognition 
\cite{tv_67,tv_150,tv_151,tv_152,tv_153}. In this 
section, we highlight some recent techniques from the CNN transformer 
pipeline domain which seek to address this issue better than transformer networks. 

In their work \cite{tv_154}, the authors study the problem 
of dense-video captioning with transformers. This requires producing 
language data for every event occurring in the video. The earlier 
techniques used for the same usually proceed sequentially: 
event-detection followed by caption-generation inside distinct 
sub-blocks. The authors of \cite{tv_154} propose a unified 
transformer architecture which learns one model for tackling both the 
aforementioned tasks jointly. Thus the proposed technique combines both 
the multi-modal MV tasks of event-detection and
caption-generation. In the first stage, a video-encoder has been used 
for obtain frame wise features, which is followed by 2 decoder units 
which propose relevant events and related captions. As a matter of fact, 
\cite{tv_154} is the first technique for dense-video 
captioning without using recurrent models. It uses self-attention based 
encoder which is fed CNN output features. Experimentation on ActivityNet 
Captions \cite{tv_155} and YouCookII 
\cite{tv_156} datasets reported valuable improvement over 
earlier RNN and double-staged techniques.

In their work \cite{tv_134}, the authors have noted in 
their work that in the multi-modal MV task learning techniques like 
VideoBERT \cite{tv_17} and ViLBERT 
\cite{tv_133} the language-processing part is generally 
kept fixed for a pre-trained model like BERT \cite{tv_3} 
for reducing the training complexity. As an alternative and also as a 
first, they have proposed PEMT, a multi-modal bidirectional transformer 
which can learn end-to-end audio-visual video data. In their model, 
short-term dependencies are first learnt using CNNs, and this is 
followed by a long-term dependency learning unit. The technique uses CNN 
features learned during its training for selection of negative samples 
which are similar to positive samples. The results obtained show that 
the concept has good implications on multi-modal task model performance.

Traditionally, CNN-based techniques for video classification usually 
performed 3D spatio-temporal manipulation on relatively small intervals 
for video understanding. In their work \cite{tv_160}, the 
authors have proposed the video transformer network (VTN) which first 
obtains frame features from a 2D CNN then applies a transformer encoder 
for learning temporal relationships. There are 2 advantages of using 
transformer encoder for the spatial features: (i) whole video is 
processed in a single pass, and (ii) training and efficiency are 
improved considerably by avoiding 3D convolution which is expensive.

\begin{figure}
\center
\includegraphics[width=10 cm]{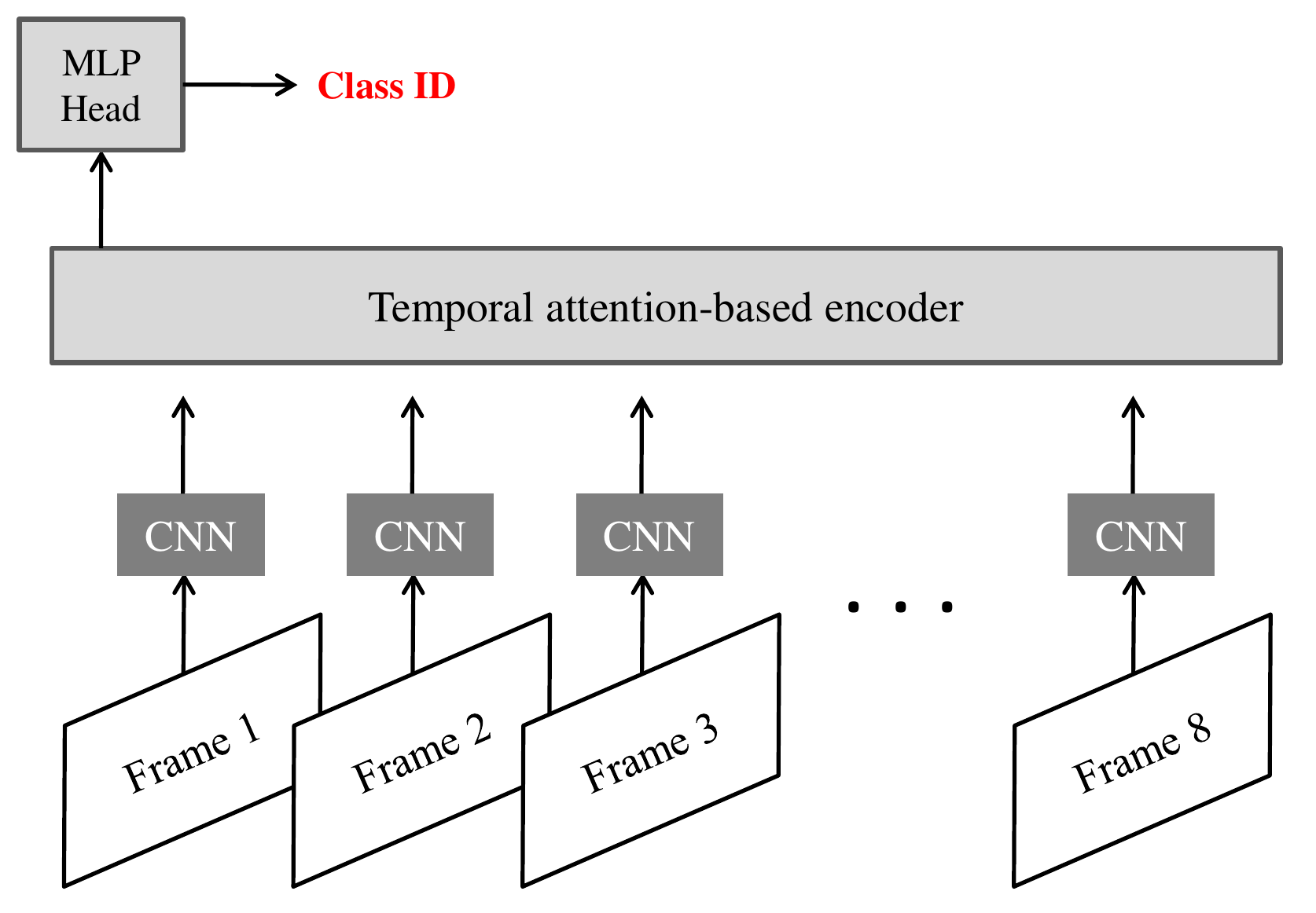}
\caption{Video transformer network (VTN) architecture \cite{tv_160}}
\label{fig6}
\center
\end{figure}

These feats make VTN suitable for learning from long videos in which 
inter-entity interactions are spread length-wise. The experiments of the 
authors on the Kinetics-400 dataset \cite{tv_67} with 
various CNN and non-CNN backbones e.g. ResNet 
\cite{tv_59}, ViT \cite{tv_11} and DeiT 
\cite{tv_12}, show good performance. \ref{fig6} shows the 
overall schematic of the proposed model.

In the next sub-section, we turn to third category of techniques of 
attention mechanisms and deep learning in machine vision, i.e. hybrid 
transformers.

\subsection{Hybrid transformers}

Transformers used to be exclusively attention based networks. However, 
some recent works have introduced two new variants i.e. convolutional 
vision transformers (CvTs) and hybrid CNN-transformer models. These 
variants are discussed below.

\subsubsection{Convolutional vision transformers}

In natural language processing (NLP) and speech recognition (SR), 
convolutional operations were used for modification of the transformer 
unit, either by changing the multi-head attention blocks with 
convolutional layers \cite{ct_38}, or by adding more 
parallel convolutional layers \cite{ct_39} or more 
sequential convolutional layers \cite{ct_13}, in order to 
capture local dependencies. Earlier research \cite{ct_37} 
proposed propagation of the attention maps to following layers by 
residual connections being transformed by convolutional operations. 

Convolutional vision transformers (CvTs) \cite{ct} improve 
the vision transformer (ViT) both in terms of performance and efficiency 
with the introduction of convolutions into ViT for yielding the best of 
both architectures. This has been achieved through 2 important 
modifications. First, a range of transformers with a novel convolutional 
token-embedding and second, a convolutional transformer unit giving 
convolutional projections. Thus they propose introduction of 
convolutional operations to 2 primary parts of the ViT viz., first, 
replacement of the 'linear projection' used for every position in the 
attention mechanism with their novel 'convolutional projection', and 
second, use of their hierarchical multistage architecture for enabling 
variable resolution of two-dimensional reshaped tokens just like CNNs. 
These fundamental changes have introduced desirable properties of CNNs 
to ViTs i.e., shift-, scale-, and distortion-invariance, while at the 
same time have maintained the merits of transformers i.e. global 
context, dynamic attention, and higher level of generalization. The 
authors validate CvT through extensive experimentation showing that 
their technique achieves state of the art performance as compared to 
other ViTs and ResNets on the ImageNet-1k dataset, with lesser 
parameters and lesser FLOPs. Also, the performance gains stay when CvT 
is pre-trained on larger datasets like ImageNet-22k 
\cite{ct_9} and is subsequently fine-tuned for downstream 
tasks. Pre-training on ImageNet-22k leads to top-1 accuracy of 87.7\% 
for the ImageNet-1k validation set. Lastly, their results demonstrate 
that positional encoding which is an important component in existing 
ViTs, can be suitably removed in CvT thus simplifying its architecture 
for higher resolution MV tasks.

\subsubsection{Hybrid CNN-transformer models}

A wide range of recent developments in handcrafted neural network models 
for machine vision tasks have asserted the important need for 
exploration of hybrid models which consist of diverse building blocks. 
At the same time, neural network model searching techniques are surging 
with expectations of reduction in human effort. In evidence brought out 
by some works \cite{hc_19,hc_61,hc_3} it is stated that 
hybrids of convolutional neural networks (CNNs) and transformers can 
perform better that both pure CNNs and pure transformers. In spite of 
this, the question that whether neural architecture search (NAS) methods 
can handle different search spaces with different candidates like CNNs 
and transformers, effectively and efficiently, leads to an open research 
area. In their work \cite{hc}, the authors propose the 
'block-wisely self-supervised neural architecture search' (BossNAS) 
which is an unsupervised NAS technique which addresses the issue of 
inaccurate model rating due to large weight-sharing space and 
supervision with bias as undertaken in earlier techniques. Going into 
specifics, they factorize the search-space into smaller blocks and also 
utilize a new self-supervision based training technique called 'ensemble 
bootstrapping', for training every block individually prior to search. 
Also, they propose a search-space called HyTra which is like a hybrid 
search-space fabric of CNNs and transformers. The fabric like 
search-space consists of model architectures similar to the common ViTs 
\cite{hc_19,hc_65,hc_13}, CNNs \cite{hc_24,hc_30} and hybrid CNN-transformers \cite{hc_61} 
at various scales. Over the same difficult search-space, their searched 
hybrid model viz. BossNet-T yields 82.2\% accuracy for ImageNet, going 
beyond EfficientNet by a margin of 2.1\% with similar computation time. 
Also, they report that their technique achieves better model rating 
accuracy on the MBConv search-space for ImageNet and on NATS-Bench size 
search-space for CIFAR-100 than the state of the art NAS techniques. The 
code and the pre-trained models are available online at 
\textcolor{red}{https://github.com/changlin31/BossNAS}.

In the next section, we discuss the major research algorithms, 
issues and trends in techniques of attention and deep learning in 
machine vision.

\section{Major research algorithms, issues and trends}

In the field of machine vision (MV), recently attention based mechanisms 
are generating a lot of interest. Pure attention based 
architectures/models are slowly and steadily proving worthy of loosening 
the grip of deep learning over MV as interesting and efficient attention 
based models continue to be built. However, pure attention based models 
come with their own set of issues. They are quite 'data-hungry' as 
they require huge amounts of data to pre-train before being able to 
be applied to MV downstream tasks after fine-tuning. As an example, 
vision transformers have to be pre-trained on the JFT dataset 
\cite{tv_39} which consists of 300 million images, and 
subsequently have to be fine-tuned on ImageNet-1K \cite{ac2_imagenet} 
before they can be used for MV tasks like image 
classification/retrieval. Also, the training times are exceedingly long 
for pre-training in transformers. Hence, reducing the 'hunger/appetite' 
of transformers is an open research area. Also, reducing the training 
time of transformers by using efficient architectures and training 
techniques is also an open research area. Reducing the computational 
load/resources for training of vision based transformers is also an open 
research area besides finding novel ways to port them to limited 
hardware/resource (portable) platforms available in the industry. A very 
large body of research work is present on deep learning and CNN based 
architectures and transformers can benefit from the same, as CNN based 
models have taken a foothold in MV. The industrial footprint of deep 
learning and CNN based models is also large. Attention based models can 
benefit from the work done and industrial footprint of deep learning 
based models. Some works \cite{hc_19,hc_61,hc_3} state 
that hybrids of convolutional neural networks (CNNs) and transformers 
can perform better that both pure CNNs and pure transformers.

Currently, the algorithms applicable to transformers benefitting from 
deep learning and CNN architecture are present in three main categories 
as discussed earlier. The first category being attention-based CNNs. The 
algorithms in this category aim to augment the performance of classical 
CNN architectures by plugging into them attention-based components/units 
in order to refine the features as and when they are used. Attention 
based CNN plugins like CBAM have been used successfully in various CNNs 
models/architectures to boost their performance at relatively small 
computational time overhead. In spite of this, the amount of attention 
available in this category is limited and the CNNs use the attention 
based mechanisms sparingly. Deeper integration and merging of attention 
based mechanisms and CNNs are required before outstanding and record 
breaking performances can be achieved. Coming to the second category of 
CNN transformer pipelines which has also been discussed, the pipeline is 
just like the earlier hybrid two-stage classifiers wherein a feature map 
generated by a 'teacher' CNN is fed to a waiting 'student' classifier 
which operates on this feature map. In this two-stage model, it is safe 
to say that the performance of the second-stage model depends on the 
image/video interpretation capability/capacity of the CNN. As such the 
architecture/design of the first-stage CNN is in question regarding its 
design-based efficacy at efficiently interpreting the image/video data. 
And it is known that there are currently a large number of CNN 
architectures available and making the correct choice is an open 
research field. Coming to the third category of Hybrid CNN-transformers, 
the merging of these two different techniques is a difficult one. 
Network architecture search (NAS) has been used to search through hybrid 
CNN-transformer search-space fabric. However, given its 
exhaustive nature requiring large computational resources and careful 
fabric design, the optimization of the same is also an open research 
area. In spite of the limitations and issues mentioned above, attention 
based mechanisms like vision transformers (ViTs) are considered having 
potential to impact the MV research and industrial body in the future. 
Combined with the power and experience of deep learning, the merger of 
the two techniques can prove to be revolutionary for both the existing 
and as well as the upcoming machine vision (MV) tasks/applications, as 
new, larger and more efficient computational hardware and software continue 
to be developed.

\section{Conclusion}

In this paper, the merger of attention based mechanisms and deep 
learning for various machine vision (MV) tasks/applications has been 
discussed. In the beginning of the paper, various types of attention 
mechanism were briefly discussed. Next, various attention based 
architectures were discussed. This was followed by discussing various 
categories of combinations of attention mechanisms and deep learning 
techniques for machine vision (MV). The various architectures and their 
associated machine vision tasks/applications were discussed. Afterwards, 
major research algorithms, issues and trends within the scope of the 
paper were discussed. By using 110+ papers as research reference in this survey, 
the readers of this paper are expected to form a 
knowledge-base and get a head-start in the area of combinational 
techniques of attention based mechanisms and deep learning for machine vision.

\textbf{Conflict of interest}

The authors declare no conflict of interest.

\bibliographystyle{spmpsci}      
\bibliography{mybibfile}   

\begin{thebibliography}{100}
\providecommand{\url}[1]{{#1}}
\providecommand{\urlprefix}{URL }
\expandafter\ifx\csname urlstyle\endcsname\relax
  \providecommand{\doi}[1]{DOI~\discretionary{}{}{}#1}\else
  \providecommand{\doi}{DOI~\discretionary{}{}{}\begingroup
  \urlstyle{rm}\Url}\fi

\bibitem{ac31_39}
\urlprefix\url{https://www.kaggle.com/c/challenges-in-representation-learning-facial-expression-recognition-challenge/data}

\bibitem{tv_39}
Revisiting the unreasonable effectiveness of data.
\newblock
  \urlprefix\url{https://ai.googleblog.com/2017/07/revisiting-unreasonable-effectiveness.html}

\bibitem{ac1_2}
Anderson, P., He, X., Buehler, C., Teney, D., Johnson, M., Gould, S., Zhang,
  L.: Bottom-up and top-down attention for image captioning and visual question
  answering.
\newblock In: 2018 IEEE/CVF Conference on Computer Vision and Pattern
  Recognition, pp. 6077--6086 (2018).
\newblock \doi{10.1109/CVPR.2018.00636}

\bibitem{tv_135}
Antol, S., Agrawal, A., Lu, J., Mitchell, M., Batra, D., Zitnick, C.L., Parikh,
  D.: Vqa: Visual question answering.
\newblock In: 2015 IEEE International Conference on Computer Vision (ICCV), pp.
  2425--2433 (2015).
\newblock \doi{10.1109/ICCV.2015.279}

\bibitem{ac1_3}
Ba, J., Mnih, V., Kavukcuoglu, K.: Multiple object recognition with visual
  attention (2015)

\bibitem{hc_3}
Bello, I.: Lambdanetworks: Modeling long-range interactions without attention.
\newblock In: International Conference on Learning Representations (2021).
\newblock \urlprefix\url{https://openreview.net/forum?id=xTJEN-ggl1b}

\bibitem{tv_160}
Berg, A., O'Connor, M., Cruz, M.T.: Keyword transformer: A self-attention model
  for keyword spotting (2021)

\bibitem{tv_13}
Carion, N., Massa, F., Synnaeve, G., Usunier, N., Kirillov, A., Zagoruyko, S.:
  End-to-end object detection with transformers.
\newblock In: A.~Vedaldi, H.~Bischof, T.~Brox, J.M. Frahm (eds.) Computer
  Vision - ECCV 2020, pp. 213--229. Springer International Publishing, Cham
  (2020)

\bibitem{tv_150}
Carreira, J., Noland, E., Hillier, C., Zisserman, A.: A short note on the
  kinetics-700 human action dataset (2019)

\bibitem{tv_19}
Chen, H., Wang, Y., Guo, T., Xu, C., Deng, Y., Liu, Z., Ma, S., Xu, C., Xu, C.,
  Gao, W.: Pre-trained image processing transformer (2020)

\bibitem{ac2_29}
Chen, L., Zhang, H., Xiao, J., Nie, L., Shao, J., Liu, W., Chua, T.S.: Sca-cnn:
  Spatial and channel-wise attention in convolutional networks for image
  captioning.
\newblock In: 2017 IEEE Conference on Computer Vision and Pattern Recognition
  (CVPR), pp. 6298--6306 (2017).
\newblock \doi{10.1109/CVPR.2017.667}

\bibitem{hc_13}
Chu, X., Tian, Z., Zhang, B., Wang, X., Wei, X., Xia, H., Shen, C.: Conditional
  positional encodings for vision transformers (2021)

\bibitem{ac2_25}
Corbetta, M., Shulman, G.L.: Control of goal-directed and stimulus-driven
  attention in the brain.
\newblock Nature reviews neuroscience \textbf{3}(3), 201--215 (2002)

\bibitem{a59}
Cornia, M., Baraldi, L., Serra, G., Cucchiara, R.: A deep multi-level network
  for saliency prediction.
\newblock In: 2016 23rd International Conference on Pattern Recognition (ICPR),
  pp. 3488--3493 (2016).
\newblock \doi{10.1109/ICPR.2016.7900174}

\bibitem{ac2_imagenet}
Deng, J., Dong, W., Socher, R., Li, L.J., Li, K., Fei-Fei, L.: Imagenet: A
  large-scale hierarchical image database.
\newblock In: 2009 IEEE Conference on Computer Vision and Pattern Recognition,
  pp. 248--255 (2009).
\newblock \doi{10.1109/CVPR.2009.5206848}

\bibitem{ct_9}
Deng, J., Dong, W., Socher, R., Li, L.J., Li, K., Fei-Fei, L.: Imagenet: A
  large-scale hierarchical image database.
\newblock In: 2009 IEEE Conference on Computer Vision and Pattern Recognition,
  pp. 248--255 (2009).
\newblock \doi{10.1109/CVPR.2009.5206848}

\bibitem{tv_3}
Devlin, J., Chang, M.W., Lee, K., Toutanova, K.: {BERT}: Pre-training of deep
  bidirectional transformers for language understanding.
\newblock In: Proceedings of the 2019 Conference of the North {A}merican
  Chapter of the Association for Computational Linguistics: Human Language
  Technologies, Volume 1 (Long and Short Papers), pp. 4171--4186. Association
  for Computational Linguistics, Minneapolis, Minnesota (2019).
\newblock \doi{10.18653/v1/N19-1423}.
\newblock \urlprefix\url{https://www.aclweb.org/anthology/N19-1423}

\bibitem{tv_25}
Doersch, C., Gupta, A., Zisserman, A.: Crosstransformers: spatially-aware
  few-shot transfer (2021)

\bibitem{hc_19}
Dosovitskiy, A., Beyer, L., Kolesnikov, A., Weissenborn, D., Zhai, X.,
  Unterthiner, T., Dehghani, M., Mindere, M., Heigold, G., Gelly, S.,
  Uszkoreit, J., Houlsby, N.: An image is worth 16x16 words: Transformers for
  image recognition at scale (2020)

\bibitem{tv_11}
Dosovitskiy, A., Beyer, L., Kolesnikov, A., Weissenborn, D., Zhai, X.,
  Unterthiner, T., Dehghani, M., Minderer, M., Heigold, G., Gelly, S.,
  Uszkoreit, J., Houlsby, N.: An image is worth 16x16 words: Transformers for
  image recognition at scale (2020)

\bibitem{sl_43}
Er, M.J., Zhang, Y., Wang, N., Pratama, M.: Attention pooling-based
  convolutional neural network for sentence modelling.
\newblock Information Sciences \textbf{373}, 388--403 (2016).
\newblock \doi{10.1016/j.ins.2016.08.084}.
\newblock
  \urlprefix\url{https://www.sciencedirect.com/science/article/pii/S0020025516306673}

\bibitem{sl_6}
Escalera, S., Bar{\'o}, X., Gonz{\`a}lez, J., Bautista, M.A., Madadi, M.,
  Reyes, M., Ponce-L{\'o}pez, V., Escalante, H.J., Shotton, J., Guyon, I.:
  Chalearn looking at people challenge 2014: Dataset and results.
\newblock In: L.~Agapito, M.M. Bronstein, C.~Rother (eds.) Computer Vision -
  ECCV 2014 Workshops, pp. 459--473. Springer International Publishing, Cham
  (2015)

\bibitem{ac2_voc}
Everingham, M., Williams, C.K.: The pascal visual object classes challenge 2007
  (voc2007) results

\bibitem{pb}
Gessert, N., Sentker, T., Madesta, F., Schmitz, R., Kniep, H., Baltruschat, I.,
  Werner, R., Schlaefer, A.: Skin lesion classification using cnns with
  patch-based attention and diagnosis-guided loss weighting.
\newblock IEEE Transactions on Biomedical Engineering \textbf{67}(2), 495--503
  (2020).
\newblock \doi{10.1109/TBME.2019.2915839}

\bibitem{tv_151}
Ging, S., Zolfaghari, M., Pirsiavash, H., Brox, T.: Coot: Cooperative
  hierarchical transformer for video-text representation learning (2020)

\bibitem{tv_18}
Girdhar, R., João~Carreira, J., Doersch, C., Zisserman, A.: Video action
  transformer network.
\newblock In: 2019 IEEE/CVF Conference on Computer Vision and Pattern
  Recognition (CVPR), pp. 244--253 (2019).
\newblock \doi{10.1109/CVPR.2019.00033}

\bibitem{dl_book}
Goodfellow, I., Bengio, Y., Courville, A.: Deep Learning.
\newblock MIT Press (2016)

\bibitem{ct_13}
Gulati, A., Qin, J., Chiu, C.C., Parmar, N., Zhang, Y., Yu, J., Han, W., Wang,
  S., Zhang, Z., Wu, Y., Pang, R.: {Conformer: Convolution-augmented
  Transformer for Speech Recognition}.
\newblock In: Proc. Interspeech 2020, pp. 5036--5040 (2020).
\newblock \doi{10.21437/Interspeech.2020-3015}.
\newblock \urlprefix\url{'http://dx.doi.org/10.21437/Interspeech.2020-3015}

\bibitem{mmir_review}
Guo, Y., Liu, Y., Georgiou, T., Lew, M.S.: A review of semantic segmentation
  using deep neural networks.
\newblock International journal of multimedia information retrieval
  \textbf{7}(2), 87--93 (2018).
\newblock \doi{10.1007/s13735-017-0141-z}

\bibitem{hafizmedical}
Hafiz, A.M., Bhat, G.M.: A survey of deep learning techniques for medical
  diagnosis.
\newblock In: M.~Tuba, S.~Akashe, A.~Joshi (eds.) Information and Communication
  Technology for Sustainable Development, pp. 161--170. Springer Singapore,
  Singapore (2020)

\bibitem{hafizmmir}
Hafiz, A.M., Bhat, G.M.: A survey on instance segmentation: state of the art.
\newblock International Journal of Multimedia Information Retrieval \textbf{9},
  171--189 (2020).
\newblock \doi{10.1007/s13735-020-00195-x}

\bibitem{hic}
Hang, R., Li, Z., Liu, Q., Ghamisi, P., Bhattacharyya, S.S.: Hyperspectral
  image classification with attention-aided cnns.
\newblock IEEE Transactions on Geoscience and Remote Sensing \textbf{59}(3),
  2281--2293 (2021).
\newblock \doi{10.1109/TGRS.2020.3007921}

\bibitem{ac2_5}
He, K., Zhang, X., Ren, S., Sun, J.: Deep residual learning for image
  recognition.
\newblock In: 2016 IEEE Conference on Computer Vision and Pattern Recognition
  (CVPR), pp. 770--778. IEEE Computer Society, Los Alamitos, CA, USA (2016).
\newblock \doi{10.1109/CVPR.2016.90}.
\newblock
  \urlprefix\url{https://doi.ieeecomputersociety.org/10.1109/CVPR.2016.90}

\bibitem{tv_59}
He, K., Zhang, X., Ren, S., Sun, J.: Deep residual learning for image
  recognition.
\newblock In: Proceedings of the IEEE Conference on Computer Vision and Pattern
  Recognition (CVPR) (2016)

\bibitem{hc_24}
He, K., Zhang, X., Ren, S., Sun, J.: Deep residual learning for image
  recognition.
\newblock In: Proceedings of the IEEE Conference on Computer Vision and Pattern
  Recognition (CVPR) (2016)

\bibitem{tv_78}
Hinton, G., Vinyals, O., Dean, J.: Distilling the knowledge in a neural network
  (2015)

\bibitem{tv_29}
Hochreiter, S., Schmidhuber, J.: {Long Short-Term Memory}.
\newblock Neural Computation \textbf{9}(8), 1735--1780 (1997).
\newblock \doi{10.1162/neco.1997.9.8.1735}.
\newblock \urlprefix\url{https://doi.org/10.1162/neco.1997.9.8.1735}

\bibitem{ac2_28}
Hu, J., Shen, L., Albanie, S., Sun, G., Wu, E.: Squeeze-and-excitation
  networks.
\newblock IEEE Transactions on Pattern Analysis and Machine Intelligence
  \textbf{42}(8), 2011--2023 (2020).
\newblock \doi{10.1109/TPAMI.2019.2913372}

\bibitem{hc_30}
Hu, J., Shen, L., Albanie, S., Sun, G., Wu, E.: Squeeze-and-excitation
  networks.
\newblock IEEE Trans. Pattern Anal. Mach. Intell. \textbf{42}(8), 2011--2023
  (2020).
\newblock \doi{10.1109/TPAMI.2019.2913372}.
\newblock \urlprefix\url{https://doi.org/10.1109/TPAMI.2019.2913372}

\bibitem{sl}
Huang, J., Zhou, W., Li, H., Li, W.: Attention-based 3d-cnns for
  large-vocabulary sign language recognition.
\newblock IEEE Transactions on Circuits and Systems for Video Technology
  \textbf{29}(9), 2822--2832 (2019).
\newblock \doi{10.1109/TCSVT.2018.2870740}

\bibitem{a52}
Huang, X., Shen, C., Boix, X., Zhao, Q.: Salicon: Reducing the semantic gap in
  saliency prediction by adapting deep neural networks.
\newblock In: 2015 IEEE International Conference on Computer Vision (ICCV), pp.
  262--270 (2015).
\newblock \doi{10.1109/ICCV.2015.38}

\bibitem{ac1_16}
Huang, X., Shen, C., Boix, X., Zhao, Q.: Salicon: Reducing the semantic gap in
  saliency prediction by adapting deep neural networks.
\newblock In: 2015 IEEE International Conference on Computer Vision (ICCV), pp.
  262--270 (2015).
\newblock \doi{10.1109/ICCV.2015.38}

\bibitem{a14}
Jetley, S., Murray, N., Vig, E.: End-to-end saliency mapping via probability
  distribution prediction.
\newblock In: 2016 IEEE Conference on Computer Vision and Pattern Recognition
  (CVPR), pp. 5753--5761 (2016).
\newblock \doi{10.1109/CVPR.2016.620}

\bibitem{tv_67}
Kay, W., Carreira, J., Simonyan, K., Zhang, B., Hillier, C., Vijayanarasimhan,
  S., Viola, F., Green, T., Back, T., Natsev, P., Suleyman, M., Zisserman, A.:
  The kinetics human action video dataset (2017)

\bibitem{tv}
Khan, S., Naseer, M., Hayat, M., Zamir, S.W., Khan, F.S., Shah, M.:
  Transformers in vision: A survey (2021)

\bibitem{tv_155}
Krishna, R., Hata, K., Ren, F., Fei-Fei, L., Niebles, J.C.: Dense-captioning
  events in videos.
\newblock In: 2017 IEEE International Conference on Computer Vision (ICCV), pp.
  706--715 (2017).
\newblock \doi{10.1109/ICCV.2017.83}

\bibitem{ac2_2}
Krizhevsky, A., et~al.: Learning multiple layers of features from tiny images
  (2009)

\bibitem{a51}
Kruthiventi, S.S.S., Ayush, K., Babu, R.V.: Deepfix: A fully convolutional
  neural network for predicting human eye fixations.
\newblock IEEE Transactions on Image Processing \textbf{26}(9), 4446--4456
  (2017).
\newblock \doi{10.1109/TIP.2017.2710620}

\bibitem{a16}
Kruthiventi, S.S.S., Gudisa, V., Dholakiya, J.H., Babu, R.V.: Saliency unified:
  A deep architecture for simultaneous eye fixation prediction and salient
  object segmentation.
\newblock In: 2016 IEEE Conference on Computer Vision and Pattern Recognition
  (CVPR), pp. 5781--5790 (2016).
\newblock \doi{10.1109/CVPR.2016.623}

\bibitem{tv_24}
Kumar, M., Weissenborn, D., Kalchbrenner, N.: Colorization transformer (2021)

\bibitem{a48}
Kümmerer, M., Theis, L., Bethge, M.: Deep gaze i: Boosting saliency prediction
  with feature maps trained on imagenet (2015)

\bibitem{a50}
Kümmerer, M., Wallis, T.S.A., Bethge, M.: Deepgaze ii: Reading fixations from
  deep features trained on object recognition (2016)

\bibitem{ac2_26}
Larochelle, H., Hinton, G.: Learning to combine foveal glimpses with a
  third-order boltzmann machine.
\newblock In: Proceedings of the 23rd International Conference on Neural
  Information Processing Systems - Volume 1, NIPS'10, pp. 1243--1251. Curran
  Associates Inc., Red Hook, NY, USA (2010)

\bibitem{dl_main}
LeCun, Y., Bengio, Y., Hinton, G.: Deep learning.
\newblock nature \textbf{521}(7553), 436--444 (2015)

\bibitem{cnn1}
Lecun, Y., Bottou, L., Bengio, Y., Haffner, P.: Gradient-based learning applied
  to document recognition.
\newblock Proceedings of the IEEE \textbf{86}(11), 2278--2324 (1998).
\newblock \doi{10.1109/5.726791}

\bibitem{cnn2}
LeCun, Y., Kavukcuoglu, K., Farabet, C.: Convolutional networks and
  applications in vision.
\newblock In: Proceedings of 2010 IEEE International Symposium on Circuits and
  Systems, pp. 253--256 (2010).
\newblock \doi{10.1109/ISCAS.2010.5537907}

\bibitem{a22}
Lee, C.Y., Xie, S., Gallagher, P., Zhang, Z., Tu, Z.: {Deeply-Supervised Nets}.
\newblock In: G.~Lebanon, S.V.N. Vishwanathan (eds.) Proceedings of the
  Eighteenth International Conference on Artificial Intelligence and
  Statistics, \emph{Proceedings of Machine Learning Research}, vol.~38, pp.
  562--570. PMLR, San Diego, California, USA (2015).
\newblock \urlprefix\url{http://proceedings.mlr.press/v38/lee15a.html}

\bibitem{tv_137}
Lee, K.H., Chen, X., Hua, G., Hu, H., He, X.: Stacked cross attention for
  image-text matching.
\newblock In: Proceedings of the European Conference on Computer Vision (ECCV)
  (2018)

\bibitem{tv_134}
Lee, S., Yu, Y., Kim, G., Breuel, T., Kautz, J., Song, Y.: Parameter efficient
  multimodal transformers for video representation learning (2020)

\bibitem{hc}
Li, C., Tang, T., Wang, G., Peng, J., Wang, B., Liang, X., Chang, X.: Bossnas:
  Exploring hybrid cnn-transformers with block-wisely self-supervised neural
  architecture search (2021)

\bibitem{a32}
Li, G., Yu, Y.: Visual saliency based on multiscale deep features.
\newblock In: 2015 IEEE Conference on Computer Vision and Pattern Recognition
  (CVPR), pp. 5455--5463 (2015).
\newblock \doi{10.1109/CVPR.2015.7299184}

\bibitem{ac31}
Li, J., Jin, K., Zhou, D., Kubota, N., Ju, Z.: Attention mechanism-based cnn
  for facial expression recognition.
\newblock Neurocomputing \textbf{411}, 340--350 (2020).
\newblock \doi{https://doi.org/10.1016/j.neucom.2020.06.014}.
\newblock
  \urlprefix\url{https://www.sciencedirect.com/science/article/pii/S0925231220309838}

\bibitem{ac1}
Li, L., Xu, M., Wang, X., Jiang, L., Liu, H.: Attention based glaucoma
  detection: A large-scale database and cnn model.
\newblock In: 2019 IEEE/CVF Conference on Computer Vision and Pattern
  Recognition (CVPR), pp. 10563--10572 (2019).
\newblock \doi{10.1109/CVPR.2019.01082}

\bibitem{ac32_4}
Li, S., Deng, W., Du, J.: Reliable crowdsourcing and deep locality-preserving
  learning for expression recognition in the wild.
\newblock In: Proceedings of the IEEE Conference on Computer Vision and Pattern
  Recognition (CVPR) (2017)

\bibitem{ac32}
Li, Y., Zeng, J., Shan, S., Chen, X.: Occlusion aware facial expression
  recognition using cnn with attention mechanism.
\newblock IEEE Transactions on Image Processing \textbf{28}(5), 2439--2450
  (2019).
\newblock \doi{10.1109/TIP.2018.2886767}

\bibitem{ac2_coco}
Lin, T.Y., Maire, M., Belongie, S., Hays, J., Perona, P., Ramanan, D.,
  Doll{\'a}r, P., Zitnick, C.L.: Microsoft coco: Common objects in context.
\newblock In: D.~Fleet, T.~Pajdla, B.~Schiele, T.~Tuytelaars (eds.) Computer
  Vision - ECCV 2014, pp. 740--755. Springer International Publishing, Cham
  (2014)

\bibitem{a4}
Liu, N., Han, J., Liu, T., Li, X.: Learning to predict eye fixations via
  multiresolution convolutional neural networks.
\newblock IEEE Transactions on Neural Networks and Learning Systems
  \textbf{29}(2), 392--404 (2018).
\newblock \doi{10.1109/TNNLS.2016.2628878}

\bibitem{tv_133}
Lu, J., Batra, D., Parikh, D., Lee, S.: Vilbert: Pretraining task-agnostic
  visiolinguistic representations for vision-and-language tasks (2019)

\bibitem{ac31_11}
Lucey, P., Cohn, J.F., Kanade, T., Saragih, J., Ambadar, Z., Matthews, I.: The
  extended cohn-kanade dataset (ck+): A complete dataset for action unit and
  emotion-specified expression.
\newblock In: 2010 IEEE Computer Society Conference on Computer Vision and
  Pattern Recognition - Workshops, pp. 94--101 (2010).
\newblock \doi{10.1109/CVPRW.2010.5543262}

\bibitem{ac31_13}
Lyons, M., Akamatsu, S., Kamachi, M., Gyoba, J.: Coding facial expressions with
  gabor wavelets.
\newblock In: Proceedings Third IEEE International Conference on Automatic Face
  and Gesture Recognition, pp. 200--205 (1998).
\newblock \doi{10.1109/AFGR.1998.670949}

\bibitem{ac32_5}
Mollahosseini, A., Hasani, B., Mahoor, M.H.: Affectnet: A database for facial
  expression, valence, and arousal computing in the wild.
\newblock IEEE Transactions on Affective Computing \textbf{10}(1), 18--31
  (2019).
\newblock \doi{10.1109/TAFFC.2017.2740923}

\bibitem{a15}
Pan, J., Sayrol, E., Giro-I-Nieto, X., McGuinness, K., O'Connor, N.E.: Shallow
  and deep convolutional networks for saliency prediction.
\newblock In: 2016 IEEE Conference on Computer Vision and Pattern Recognition
  (CVPR), pp. 598--606 (2016).
\newblock \doi{10.1109/CVPR.2016.71}

\bibitem{a61}
Pinheiro, P.O., Lin, T.Y., Collobert, R., Doll{\'a}r, P.: Learning to refine
  object segments.
\newblock In: B.~Leibe, J.~Matas, N.~Sebe, M.~Welling (eds.) Computer Vision -
  ECCV 2016, pp. 75--91. Springer International Publishing, Cham (2016)

\bibitem{tv_79}
Radosavovic, I., Kosaraju, R.P., Girshick, R., He, K., Dollar, P.: Designing
  network design spaces.
\newblock In: 2020 IEEE/CVF Conference on Computer Vision and Pattern
  Recognition (CVPR), pp. 10425--10433 (2020).
\newblock \doi{10.1109/CVPR42600.2020.01044}

\bibitem{tv_20}
Ramesh, A., Pavlov, M., Goh, G., Gray, S., Chen, M., Child, R., Misra, V.,
  Mishkin, P., Krueger, G., Agarwal, S., et~al.: Dall{\textperiodcentered} e:
  Creating images from text.
\newblock OpenAI blog. https://openai. com/blog/dall-e  (2021)

\bibitem{ac1_28}
Ren, S., He, K., Girshick, R., Sun, J.: Faster r-cnn: Towards real-time object
  detection with region proposal networks.
\newblock In: Proceedings of the 28th International Conference on Neural
  Information Processing Systems - Volume 1, NIPS'15, pp. 91--99. MIT Press,
  Cambridge, MA, USA (2015)

\bibitem{tv_83}
Ren, S., He, K., Girshick, R., Sun, J.: Faster r-cnn: Towards real-time object
  detection with region proposal networks.
\newblock IEEE Transactions on Pattern Analysis and Machine Intelligence
  \textbf{39}(6), 1137--1149 (2017).
\newblock \doi{10.1109/TPAMI.2016.2577031}

\bibitem{ac2_24}
Rensink, R.A.: The dynamic representation of scenes.
\newblock Visual Cognition \textbf{7}(1-3), 17--42 (2000).
\newblock \doi{10.1080/135062800394667}.
\newblock \urlprefix\url{https://doi.org/10.1080/135062800394667}

\bibitem{ac2_18}
Selvaraju, R.R., Cogswell, M., Das, A., Vedantam, R., Parikh, D., Batra, D.:
  Grad-cam: Visual explanations from deep networks via gradient-based
  localization.
\newblock In: 2017 IEEE International Conference on Computer Vision (ICCV), pp.
  618--626 (2017).
\newblock \doi{10.1109/ICCV.2017.74}

\bibitem{tv_152}
Seong, H., Hyun, J., Kim, E.: Video multitask transformer network.
\newblock In: 2019 IEEE/CVF International Conference on Computer Vision
  Workshop (ICCVW), pp. 1553--1561 (2019).
\newblock \doi{10.1109/ICCVW.2019.00194}

\bibitem{ac1_30}
Sharma, S., Kiros, R., Salakhutdinov, R.: Action recognition using visual
  attention.
\newblock In: International Conference on Learning Representations (ICLR)
  Workshop (2016).
\newblock \urlprefix\url{https://arxiv.org/abs/1511.04119}

\bibitem{dl_review}
Shrestha, A., Mahmood, A.: Review of deep learning algorithms and
  architectures.
\newblock IEEE Access \textbf{7}, 53040--53065 (2019).
\newblock \doi{10.1109/ACCESS.2019.2912200}

\bibitem{ac1_31}
Simonyan, K., Zisserman, A.: Very deep convolutional networks for large-scale
  image recognition (2015)

\bibitem{hc_61}
Srinivas, A., Lin, T.Y., Parmar, N., Shlens, J., Abbeel, P., Vaswani, A.:
  Bottleneck transformers for visual recognition (2021)

\bibitem{tv_22}
Su, W., Zhu, X., Cao, Y., Li, B., Lu, L., Wei, F., Dai, J.: Vl-bert:
  Pre-training of generic visual-linguistic representations.
\newblock In: International Conference on Learning Representations (2020).
\newblock \urlprefix\url{https://openreview.net/forum?id=SygXPaEYvH}

\bibitem{tv_17}
Sun, C., Myers, A., Vondrick, C., Murphy, K., Schmid, C.: Videobert: A joint
  model for video and language representation learning.
\newblock In: 2019 IEEE/CVF International Conference on Computer Vision (ICCV),
  pp. 7463--7472 (2019).
\newblock \doi{10.1109/ICCV.2019.00756}

\bibitem{tv_21}
Tan, H., Bansal, M.: {LXMERT}: Learning cross-modality encoder representations
  from transformers.
\newblock In: Proceedings of the 2019 Conference on Empirical Methods in
  Natural Language Processing and the 9th International Joint Conference on
  Natural Language Processing (EMNLP-IJCNLP), pp. 5100--5111. Association for
  Computational Linguistics, Hong Kong, China (2019).
\newblock \doi{10.18653/v1/D19-1514}.
\newblock \urlprefix\url{https://www.aclweb.org/anthology/D19-1514}

\bibitem{tv_80}
Tan, M., Le, Q.: {E}fficient{N}et: Rethinking model scaling for convolutional
  neural networks.
\newblock In: K.~Chaudhuri, R.~Salakhutdinov (eds.) Proceedings of the 36th
  International Conference on Machine Learning, \emph{Proceedings of Machine
  Learning Research}, vol.~97, pp. 6105--6114. PMLR (2019).
\newblock \urlprefix\url{http://proceedings.mlr.press/v97/tan19a.html}

\bibitem{ac4}
Tian, C., Xu, Y., Li, Z., Zuo, W., Fei, L., Liu, H.: Attention-guided cnn for
  image denoising.
\newblock Neural Networks \textbf{124}, 117--129 (2020).
\newblock \doi{https://doi.org/10.1016/j.neunet.2019.12.024}.
\newblock
  \urlprefix\url{https://www.sciencedirect.com/science/article/pii/S0893608019304241}

\bibitem{tv_12}
Touvron, H., Cord, M., Douze, M., Massa, F., Sablayrolles, A., Jégou, H.:
  Training data-efficient image transformers \& distillation through attention
  (2021)

\bibitem{hc_65}
Touvron, H., Cord, M., Douze, M., Massa, F., Sablayrolles, A., Jégou, H.:
  Training data-efficient image transformers \& distillation through attention
  (2021)

\bibitem{sl_18}
Tran, D., Bourdev, L., Fergus, R., Torresani, L., Paluri, M.: Learning
  spatiotemporal features with 3d convolutional networks.
\newblock In: Proceedings of the IEEE International Conference on Computer
  Vision (ICCV) (2015)

\bibitem{tv_1}
Vaswani, A., Shazeer, N., Parmar, N., Uszkoreit, J., Jones, L., Gomez, A.N.,
  Kaiser, u., Polosukhin, I.: Attention is all you need.
\newblock In: Proceedings of the 31st International Conference on Neural
  Information Processing Systems, NIPS'17, pp. 6000--6010. Curran Associates
  Inc., Red Hook, NY, USA (2017)

\bibitem{tv_138}
Vinyals, O., Toshev, A., Bengio, S., Erhan, D.: Show and tell: A neural image
  caption generator.
\newblock In: 2015 IEEE Conference on Computer Vision and Pattern Recognition
  (CVPR), pp. 3156--3164 (2015).
\newblock \doi{10.1109/CVPR.2015.7298935}

\bibitem{ac2_27}
Wang, F., Jiang, M., Qian, C., Yang, S., Li, C., Zhang, H., Wang, X., Tang, X.:
  Residual attention network for image classification.
\newblock In: 2017 IEEE Conference on Computer Vision and Pattern Recognition
  (CVPR), pp. 6450--6458 (2017).
\newblock \doi{10.1109/CVPR.2017.683}

\bibitem{DeepVisualAttentionPred}
Wang, W., Shen, J.: Deep visual attention prediction.
\newblock IEEE Transactions on Image Processing \textbf{27}(5), 2368--2378
  (2018).
\newblock \doi{10.1109/TIP.2017.2787612}

\bibitem{tv_23}
Wang, X., Yeshwanth, C., Niebner, M.: Sceneformer: Indoor scene generation with
  transformers (2021)

\bibitem{tv_153}
Wang, Y., Xu, Z., Wang, X., Shen, C., Cheng, B., Shen, H., Xia, H.: End-to-end
  video instance segmentation with transformers (2021)

\bibitem{ct_37}
Wang, Y., Yang, Y., Bai, J., Zhang, M., Bai, J., Yu, J., Zhang, C., Huang, G.,
  Tong, Y.: Evolving attention with residual convolutions (2021)

\bibitem{ac2_30}
Woo, S., Hwang, S., Kweon, I.S.: Stairnet: Top-down semantic aggregation for
  accurate one shot detection.
\newblock In: 2018 IEEE Winter Conference on Applications of Computer Vision
  (WACV), pp. 1093--1102 (2018).
\newblock \doi{10.1109/WACV.2018.00125}

\bibitem{ac2}
Woo, S., Park, J., Lee, J.Y., Kweon, I.S.: Cbam: Convolutional block attention
  module.
\newblock In: V.~Ferrari, M.~Hebert, C.~Sminchisescu, Y.~Weiss (eds.) Computer
  Vision - ECCV 2018, pp. 3--19. Springer International Publishing, Cham (2018)

\bibitem{ct_38}
Wu, F., Fan, A., Baevski, A., Dauphin, Y.N., Auli, M.: Pay less attention with
  lightweight and dynamic convolutions (2019)

\bibitem{ct}
Wu, H., Xiao, B., Codella, N., Liu, M., Dai, X., Yuan, L., Zhang, L.: Cvt:
  Introducing convolutions to vision transformers (2021)

\bibitem{ct_39}
Wu, Z., Liu, Z., Lin, J., Lin, Y., Han, S.: Lite transformer with long-short
  range attention (2020)

\bibitem{a58}
Xie, S., Tu, Z.: Holistically-nested edge detection.
\newblock In: 2015 IEEE International Conference on Computer Vision (ICCV), pp.
  1395--1403 (2015).
\newblock \doi{10.1109/ICCV.2015.164}

\bibitem{ac1_35}
Xu, K., Ba, J., Kiros, R., Cho, K., Courville, A., Salakhudinov, R., Zemel, R.,
  Bengio, Y.: Show, attend and tell: Neural image caption generation with
  visual attention.
\newblock In: F.~Bach, D.~Blei (eds.) Proceedings of the 32nd International
  Conference on Machine Learning, \emph{Proceedings of Machine Learning
  Research}, vol.~37, pp. 2048--2057. PMLR, Lille, France (2015).
\newblock \urlprefix\url{http://proceedings.mlr.press/v37/xuc15.html}

\bibitem{ac1_37}
Xu, M., Li, C., Liu, Y., Deng, X., Lu, J.: A subjective visual quality
  assessment method of panoramic videos.
\newblock In: 2017 IEEE International Conference on Multimedia and Expo (ICME),
  pp. 517--522 (2017).
\newblock \doi{10.1109/ICME.2017.8019351}

\bibitem{a28}
Yang, C., Zhang, L., Lu, H., Ruan, X., Yang, M.H.: Saliency detection via
  graph-based manifold ranking.
\newblock In: 2013 IEEE Conference on Computer Vision and Pattern Recognition,
  pp. 3166--3173 (2013).
\newblock \doi{10.1109/CVPR.2013.407}

\bibitem{tv_16}
Yang, F., Yang, H., Fu, J., Lu, H., Guo, B.: Learning texture transformer
  network for image super-resolution.
\newblock In: 2020 IEEE/CVF Conference on Computer Vision and Pattern
  Recognition (CVPR), pp. 5790--5799 (2020).
\newblock \doi{10.1109/CVPR42600.2020.00583}

\bibitem{a60}
Yao, X., Han, J., Zhang, D., Nie, F.: Revisiting co-saliency detection: A novel
  approach based on two-stage multi-view spectral rotation co-clustering.
\newblock IEEE Transactions on Image Processing \textbf{26}(7), 3196--3209
  (2017).
\newblock \doi{10.1109/TIP.2017.2694222}

\bibitem{tv_26}
Ye, H.J., Hu, H., Zhan, D.C., Sha, F.: Few-shot learning via embedding
  adaptation with set-to-set functions.
\newblock In: 2020 IEEE/CVF Conference on Computer Vision and Pattern
  Recognition (CVPR), pp. 8805--8814 (2020).
\newblock \doi{10.1109/CVPR42600.2020.00883}

\bibitem{tv_15}
Ye, L., Rochan, M., Liu, Z., Wang, Y.: Cross-modal self-attention network for
  referring image segmentation.
\newblock In: 2019 IEEE/CVF Conference on Computer Vision and Pattern
  Recognition (CVPR), pp. 10494--10503 (2019).
\newblock \doi{10.1109/CVPR.2019.01075}

\bibitem{ac1_39}
Yu, Y., Choi, J., Kim, Y., Yoo, K., Lee, S.H., Kim, G.: Supervising neural
  attention models for video captioning by human gaze data.
\newblock In: 2017 IEEE Conference on Computer Vision and Pattern Recognition
  (CVPR), pp. 6119--6127 (2017).
\newblock \doi{10.1109/CVPR.2017.648}

\bibitem{tv_136}
Zellers, R., Bisk, Y., Farhadi, A., Choi, Y.: From recognition to cognition:
  Visual commonsense reasoning.
\newblock In: 2019 IEEE/CVF Conference on Computer Vision and Pattern
  Recognition (CVPR), pp. 6713--6724 (2019).
\newblock \doi{10.1109/CVPR.2019.00688}

\bibitem{ac31_25}
Zhao, G., Huang, X., Taini, M., Li, S.Z., Pietikäinen, M.: Facial expression
  recognition from near-infrared videos.
\newblock Image and Vision Computing \textbf{29}(9), 607--619 (2011).
\newblock \doi{https://doi.org/10.1016/j.imavis.2011.07.002}.
\newblock
  \urlprefix\url{https://www.sciencedirect.com/science/article/pii/S0262885611000515}

\bibitem{a57}
Zhao, R., Ouyang, W., Li, H., Wang, X.: Saliency detection by multi-context
  deep learning.
\newblock In: 2015 IEEE Conference on Computer Vision and Pattern Recognition
  (CVPR), pp. 1265--1274 (2015).
\newblock \doi{10.1109/CVPR.2015.7298731}

\bibitem{tv_156}
Zhou, L., Xu, C., Corso, J.: Towards automatic learning of procedures from web
  instructional videos.
\newblock Proceedings of the AAAI Conference on Artificial Intelligence
  \textbf{32}(1) (2018).
\newblock
  \urlprefix\url{https://ojs.aaai.org/index.php/AAAI/article/view/12342}

\bibitem{tv_154}
Zhou, L., Zhou, Y., Corso, J.J., Socher, R., Xiong, C.: End-to-end dense video
  captioning with masked transformer.
\newblock In: 2018 IEEE/CVF Conference on Computer Vision and Pattern
  Recognition, pp. 8739--8748 (2018).
\newblock \doi{10.1109/CVPR.2018.00911}

\bibitem{tv_14}
Zhu, X., Su, W., Lu, L., Li, B., Wang, X., Dai, J.: Deformable detr: Deformable
  transformers for end-to-end object detection (2021)

\end{thebibliography}

\end{document}